%% file: egpaper_final.tex
\documentclass[10pt,twocolumn,letterpaper]{article}

\usepackage{iccv}
\usepackage{times}
\usepackage{epsfig}
\usepackage{graphicx}
\usepackage{amsmath}
\usepackage{amssymb}
\usepackage{kotex}
\usepackage[usenames,dvipsnames]{xcolor}
\usepackage{booktabs}
\usepackage{multirow}
\usepackage{soul}
\usepackage{multicol}
\usepackage[pagebackref=true,breaklinks=true,letterpaper=true,colorlinks,bookmarks=false]{hyperref}

\iccvfinalcopy % *** Uncomment this line for the final submission

\def\iccvPaperID{****} % *** Enter the ICCV Paper ID here

\begin{document}

%%%%%%%%% TITLE
\title{A Comprehensive Overhaul of Feature Distillation}

\author{
  Byeongho Heo\textsuperscript{1,3}\thanks{This work was done when authors were in research internship at \textit{Clova AI Research}, NAVER corp.}
  \,\,
  Jeesoo Kim\textsuperscript{2}\footnotemark[1]
  \,\,
  Sangdoo Yun\textsuperscript{1}
  \,\,
  Hyojin Park\textsuperscript{2}\footnotemark[1]
  \,\,
  Nojun Kwak\textsuperscript{2}
  \,\,
  Jin Young Choi\textsuperscript{3}
  \\
  \small{\texttt{\{bhheo, kimjiss0305, wolfrun, nojunk, jychoi\}@snu.ac.kr},\; 
  \texttt{sangdoo.yun@navercorp.com}}\\
  \textsuperscript{1}Clova AI Research, NAVER Corp, Korea\\
  \textsuperscript{2}GSCST, Seoul National University, Korea\\
  \textsuperscript{3}Department of ECE, ASRI, Seoul National University, Korea\\
}

\maketitle
%\thispagestyle{empty}

%%%%%%%%% ABSTRACT
\begin{abstract}
We investigate the design aspects of feature distillation methods achieving network compression and propose a novel feature distillation method in which the distillation loss is designed to make a synergy among various aspects: teacher transform, student transform, distillation feature position and distance function. Our proposed distillation loss includes a feature transform with a newly designed margin ReLU, a new distillation feature position, and a partial $L_2$ distance function to skip redundant information giving adverse effects to the compression of student.
In ImageNet, our proposed method achieves 21.65\% of top-1 error with ResNet50, which outperforms the performance of the teacher network, ResNet152.
Our proposed method is evaluated on various tasks such as image classification, object detection and semantic segmentation and achieves a significant performance improvement in all tasks.
The code is available at \href{https://sites.google.com/view/byeongho-heo/overhaul}{bhheo.github.io/overhaul}
\end{abstract}

\input{1_Introduction.tex}

\input{2_Preliminary.tex}
\input{3_Proposed.tex}
\input{4_Experiments.tex}
\input{5_Conclusion.tex}

\section*{Acknowledgement}
This work was supported by Next-Generation ICD Program through NRF funded by Ministry of S\&ICT [2017M3C4A7077582] and ICT R\&D program MSIP/IITP [2017-0-00306, Outdoor Surveillance Robots],  
We appreciate support of Clova AI members, especially Nigel Fernandez for proofreading the manuscript, Dongyoon Han for
providing help on implementation and Jung-Woo Ha for insightful comments.
We also thank the NSML Team for providing an excellent experiment platform.

{\small
\bibliographystyle{ieee_fullname}
\bibliography{egbib}
}

\clearpage
\input{Appendix.tex}

\end{document}

%% file: 1_Introduction.tex
\section{Introduction}

Experiencing remarkable advances in many machine learning tasks using neural networks, researchers have started to work on network compression and enhancement.
Several approaches such as model pruning, model quantization and knowledge distillation have been suggested to make the model smaller and cost-efficient.
Among them, \textit{knowledge distillation} is being actively investigated.
Knowledge distillation refers to the method that helps the training process of a smaller network (student) under the supervision of a larger network (teacher).
Unlike other compression methods, it can downsize a network regardless of the structural difference between the teacher and the student network.
Allowing architectural flexibility, knowledge distillation is emerging as a next generation approach of network compression.

Hinton \textit{et al.}~\cite{hinton2015distilling} proposed a knowledge distillation (KD) method using the softmax output of the teacher network.
This method can be applied to any pair of network architectures since the dimensions of both outputs are the same. 
However, the output of a high-performance teacher network is not significantly different from the ground truth.
Thus, transferring only the output is similar to training the student with the ground truth, making the performance of output distillation limited. % does not contribute much to the performance improvement of the student network.
To make better use of the information contained in the teacher network, several approaches have been proposed for feature distillation instead of output distillation.
FitNets~\cite{romero2014fitnets} have proposed a method that encourages a student network to mimic the hidden feature values of a teacher network.
Although feature distillation was a promising approach, the performance improvement by
% Feature distillation had been a new method with a huge potential, but the performance of 
FitNets was not significant.

\begin{figure}[t]
	\centering
    \includegraphics[width=\columnwidth]{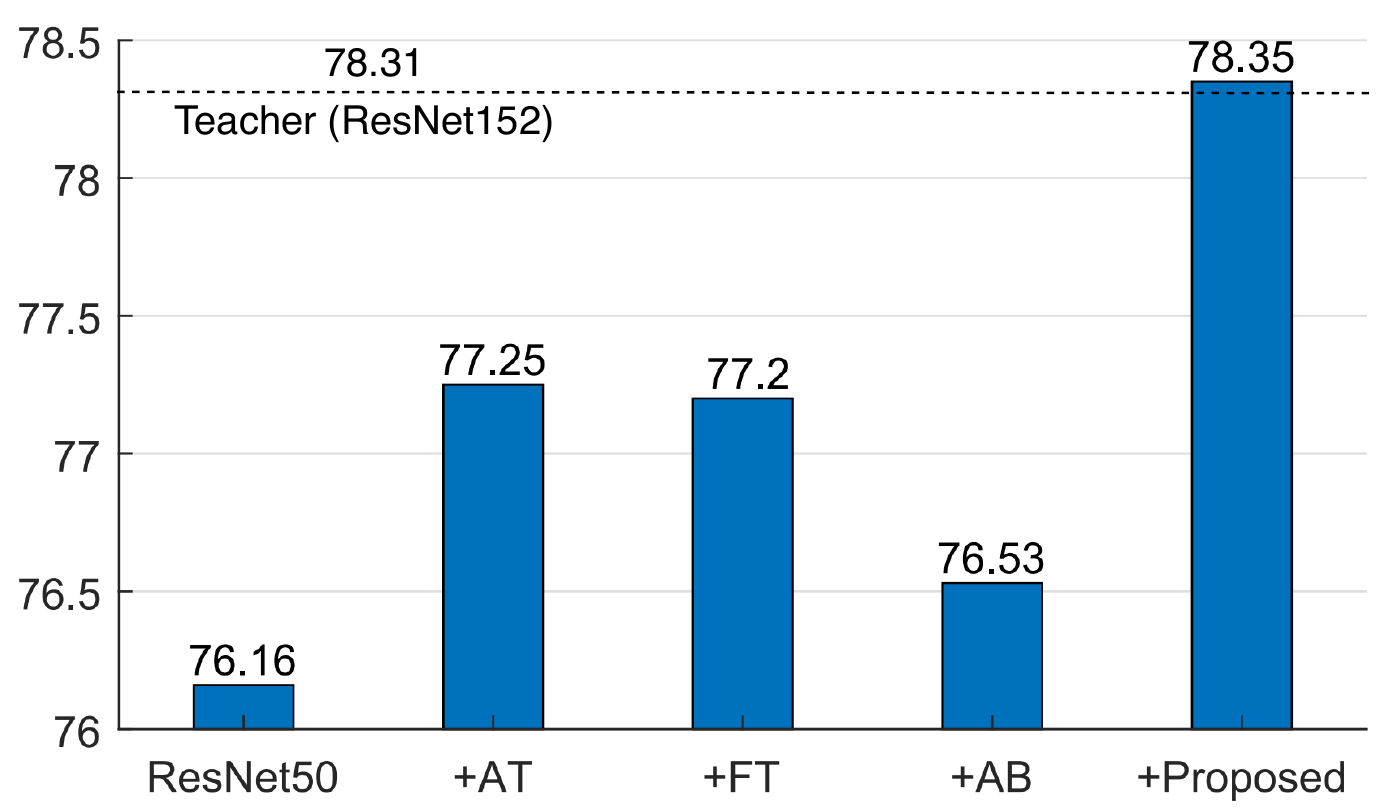}
  \caption{Performance of distillation methods: AT~\cite{zagoruyko2016paying}, FT~\cite{kim2018paraphrasing}, AB~\cite{heo2018knowledge} and proposed method on ImageNet. The graph shows accuracy(\%) of ResNet50 trained with each distillation method. Note that ResNet152 with 78.31\% accuracy is used as a teacher.}
\label{fig:head_performance}
\vspace{-0.2cm}
\end{figure}

After FitNets, variant methods of feature distillation have been proposed as follows.
The methods in \cite{zagoruyko2016paying,yim2017gift} transform the feature into a representation having a reduced dimension and transfer it to the student.
In spite of the reduced dimension, it has been reported that the abstracted feature representation does lead to an improved performance.
Recent methods (FT~\cite{kim2018paraphrasing}, AB~\cite{heo2018knowledge}) have been proposed to increase the amount of transferred information in distillation.
FT \cite{kim2018paraphrasing} encodes the feature into a `factor' using an auto-encoder to alleviate the leakage of information.
AB \cite{heo2018knowledge} focuses on activation of a network with only the sign of features being transferred.
Both methods show a better distillation performance by increasing the amount of transferred information.
However, FT~\cite{kim2018paraphrasing} and AB~\cite{heo2018knowledge} deform feature values of the teacher, which leaves a further room for the performance to be improved.

In this paper, we further improve the performance of feature distillation by proposing a new feature distillation loss which is designed via investigation of various design aspects: teacher transform, student transform, distillation feature position and distance function. 
Our method aims to transfer two factors from features.
The first target is the magnitude of feature response after ReLU, since it carries most of the feature information.
The second is the activation status of each neuron. 
Recent studies~\cite{pan2016expressiveness,heo2018knowledge} have shown that the activation of neurons strongly represents the expressiveness of a network, and it should be considered in distillation.
To this purpose, we propose a margin ReLU function, change the distillation feature position to the front of ReLU, and use a partial $L_2$ distance function to skip the distillation of unnecessary information.
The proposed loss significantly improves performance of feature distillation.
In our experiments, we have evaluated our proposed method in various domains including classification (CIFAR~\cite{CIFAR}, ImageNet~\cite{ImageNet}), object detection (PASCAL VOC~\cite{Everingham2015VOC}) and semantic segmentation (PASCAL VOC).
As shown in Fig.~\ref{fig:head_performance}, in our experiments, the proposed method shows a performance superior to the existing state-of-the-art methods and even the teacher model.

%% file: 2_Preliminary.tex
\input{Tables/table_comparison.tex}

\begin{figure}[t]
	\centering
    \includegraphics[width=\columnwidth]{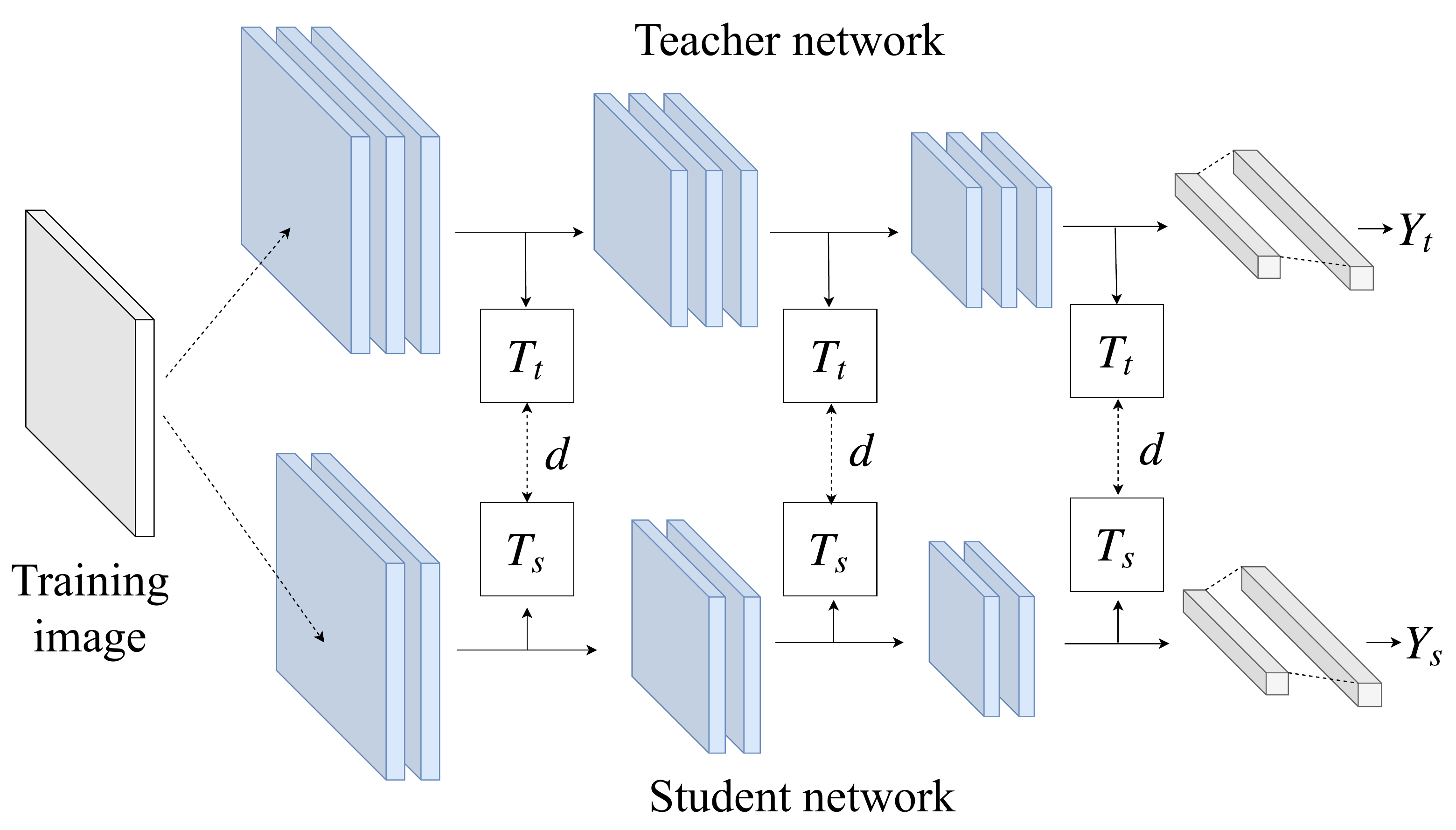}
  \caption{The general training scheme of feature distillation. The form of teacher transform $T_{t}$, student transform $T_{s}$ and distance $d$ differ from method to method.}
\label{fig:featuredistillation}
\vspace{-0.2cm}
\end{figure}

\section{Motivation}
\label{section:preliminary}

In this section, we investigate the design aspects of feature distillation methods achieving network compression and present novel aspects of our approach distinctive to the preceding methods.
First, we describe a general form of loss function in feature distillation.
As shown in Fig.~\ref{fig:featuredistillation}, the feature of the teacher network is denoted as $\boldsymbol{F}_{t}$ and the feature of the student network is $\boldsymbol{F}_{s}$.
To match the feature dimension, $T_{t}$ and $T_{s}$ respectively, we transform the feature $\boldsymbol{F}_{t}$ and $\boldsymbol{F}_{s}$.
A distance $d$ between the transformed features is used as a loss function $L_{distill}$.
In other words, the loss function of feature distillation is generalized as
\begin{equation}
  L_{distill} = d(T_{t}(\boldsymbol{F}_{t}), T_{s}(\boldsymbol{F}_{s})).
  \label{equ:fdistobjective}
\end{equation}
The student network is trained by minimizing the distillation loss $L_{distill}$. 

It is desirable to design the distillation loss so as to transfer all feature information without missing any important information from the teacher. 
To achieve this, we aim to design a new feature distillation loss in which all important teacher's information is  transferred as much as possible to improve the distillation performance. 
To get an idea for this purpose, we analyze the design aspects of feature distillation loss.
As described in Table~\ref{table:methodcomparison}, the design aspects of feature distillation loss are categorized into 4 categories: teacher transform, student transform, distillation feature position and distance function.

\textbf{Teacher transform.} 
A teacher transform $T_t$ converts the teacher's hidden features into an easy-to-transfer form.
It is an important part of feature distillation and also a main cause of the information missing in distillation.
AT~\cite{zagoruyko2016paying}, FSP~\cite{yim2017gift} and Jacobian~\cite{suraj2018jacobian} reduce the dimension of the feature vector via the teacher transform which causes serious information missing.
FT~\cite{kim2018paraphrasing} uses a compression ratio determined by the user and AB~\cite{heo2018knowledge} utilizes the original feature in the form of binarized values, making both methods to use features different from the original ones.
Except FitNets~\cite{romero2014fitnets}, most teacher transforms of existing approaches cause an information missing in the teacher's feature used in the distillation loss.
Since features include both adverse and beneficial information, it is important to distinguish them and avoid missing the beneficial information.
In the proposed method, we use a new ReLU activation, called margin ReLU, for the teacher transform. 
In our margin ReLU, the positive (beneficial) information is used without any transformation while the negative (adverse) information is suppressed.
As a result, the proposed method can perform distillation without missing the beneficial information.

\textbf{Student transform.}
Typically, the student transform $T_s$ uses the same function as the teacher transform $T_t$.
Therefore, in methods like AT~\cite{zagoruyko2016paying}, FSP~\cite{yim2017gift}, Jacobian~\cite{suraj2018jacobian} and FT~\cite{kim2018paraphrasing}, the same amount of information is lost in both the teacher transform and the student transform.
FitNets and AB do not reduce the dimension of teacher's feature 
and use a 1$\times$1 convolutional layer as a student transform to match the feature dimension with the teacher.
In this case, the feature size of the student does not decrease, but rather increases, so there is no information missing.
In our method, we use this asymmetric format of transformations as the student transform.

\textbf{Distillation feature position.}
Besides the types of feature transformation, we should be careful in picking the location in which distillation occurs.
FitNets uses the end of an arbitrary middle layer as the distillation point, which has been shown to have a poor performance.
We refer to a group of layers with the same spatial size as a layer group~\cite{Zagoruyko2016WRN,Han2017pyramidnet}.
In AT~\cite{zagoruyko2016paying}, FSP~\cite{yim2017gift} and Jacobian~\cite{suraj2018jacobian}, the distillation point lies at the end of each layer group, whereas in FT it lies at the end of only the last layer group.
This has led to better results than FitNets but still lacks consideration about the ReLU-activated values of the teacher. 
ReLU allows the beneficial information (positive) to pass through and filters out the adverse information (negative).
Therefore, knowledge distillation must be designed under the acknowledgement of this information dissolution.
In our method, we design the distillation loss to bring the features in front of the ReLU function, called pre-ReLU.
Positive and negative values are preserved in the pre-ReLU position without any deformation.
So, it is suitable for distillation.

\textbf{Distance function}.
Most distillation methods naively adopt $L_2$ or $L_1$ distance.
However, in our method, we need to design an appropriate distance function according to our teacher transform, and our distillation point in the pre-ReLU position. 
In our design, the pre-ReLU information is transferred from teacher to student, but negative values of the pre-ReLU feature contain adverse information.
The negative values of the pre-ReLU feature are blocked by the ReLU activation and not used by the teacher network.
The transfer of all values may have an adverse effect to the student network. 
To handle this issue,
we propose a new distance function, called partial $L_2$ distance, which is designed to skip the distillation of information on a negative region.

%% file: Tables/table_comparison.tex
\begin{table*}[t]
\begin{center}
\begin{tabular}{@{}lccccc@{}}
\toprule
Method   & \begin{tabular}[c]{@{}c@{}}Teacher\\ transform\end{tabular} & \begin{tabular}[c]{@{}c@{}}Student\\ transform\end{tabular} & \begin{tabular}[c]{@{}c@{}}Distillation\\ feature position\end{tabular} &
Distance &\begin{tabular}[c]{@{}c@{}}Missing\\information\end{tabular} \\ \midrule
FitNets~\cite{romero2014fitnets}   & None  & 1$\times$1 conv    & Mid layer       & $L_2$   & None          \\
AT~\cite{zagoruyko2016paying}       & Attention     & Attention         & End of group  &  $L_2$        & Channel dims  \\
FSP~\cite{yim2017gift}      & Correlation   & Correlation       & End of group        & $L_2$  & Spatial dims  \\
Jacobian~\cite{suraj2018jacobian} & Gradient      & Gradient          & End of group      & $L_2$   & Channel dims  \\
FT~\cite{kim2018paraphrasing}       & Auto-encoder  & Auto-encoder     & End of group     & $L_1$   & Auto-encoded  \\
AB~\cite{heo2018knowledge}       & Binarization  & 1$\times$1 conv        & Pre-ReLU          & Marginal $L_2$    & Feature values         \\ \midrule
Proposed & Margin ReLU   & 1$\times$1 conv         & Pre-ReLU & Partial $L_2$  & Negative features  \\ \bottomrule
\end{tabular}
\end{center}
\caption{Difference in various kinds of feature distillation. Most distillation use teacher transform with information loss.}
\label{table:methodcomparison}
\vspace{-0.2cm}
\end{table*}

%% file: 3_Proposed.tex
\section{Approach}

In this section, we describe our distillation method outlined in section 2.
We first describe the location where the distillation occurs in our method and then explain about the newly designed loss function.

\begin{figure}[t]
	\centering
    \includegraphics[width=\columnwidth]{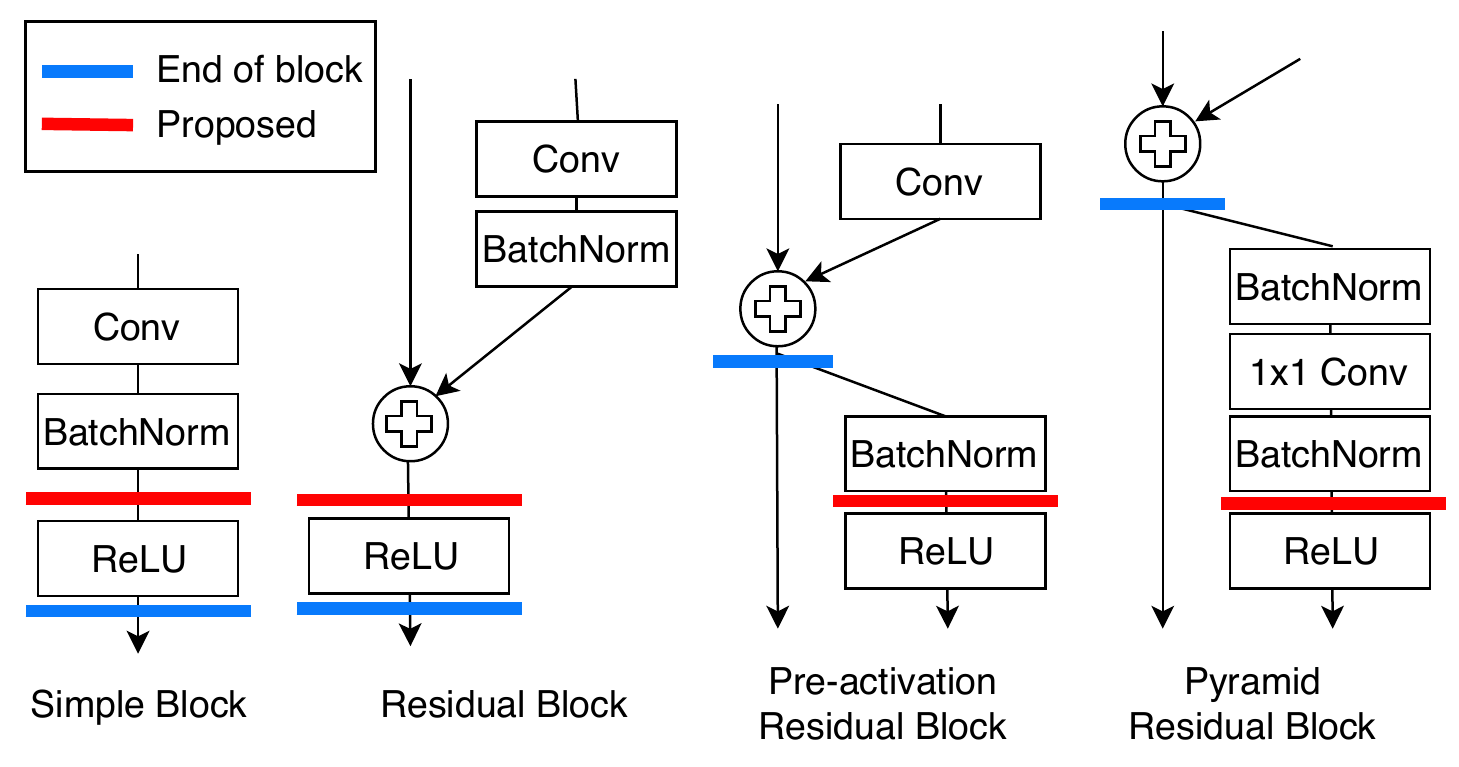}
  \caption{Position of distillation target layer. We place the distillation layer between the last block and the first ReLU. The exact location differs according to the network architecture.}
\label{fig:position}
\vspace{-0.2cm}
\end{figure}

\subsection{Distillation position}
\label{sec:distill_position}

The activation function is a crucial component of neural networks.
The non-linearity of a neural network attributes to this function.
The performance of the model is significantly influenced by the type of activation function.
Among various activation functions, rectified linear unit (ReLU)~\cite{Nair2010ReLU} is used in most computer vision tasks.
Most networks~\cite{alexnet,Simonyan2014vgg,googlenet,resnet,He2016preact,Zagoruyko2016WRN,Han2017pyramidnet,andrew2017mobilenet,Sandler2018mobileV2} use ReLU or modified versions very similar to ReLU~\cite{Maas2013leakyrelu,Krizhevsky2010relu6}.
ReLU simply applies a linear mapping for positive values.
For negative values, it eliminates the values and fixes them to zero, which prevents the unnecessary information from going backward.
With a careful design of knowledge distillation considering ReLU, it is possible to transfer only the necessary information.
Unfortunately, most of the preceding research don't take a serious consideration of the activation function.
We define the minimum unit of the network, such as the residual block in ResNet~\cite{resnet} and the Conv-ReLU in VGG~\cite{Simonyan2014vgg}, as a layer block.
The distillation in most methods occurs at the end of the layer block ignoring whether it is related to ReLU or not.

In our proposed method, the position of the distillation lies between the first ReLU and the end of layer block.
This positioning enables the student to reach the preserved information of the teacher before it passes through ReLU.
Fig.~\ref{fig:position} depicts the distillation position of some architectures.
In case of the simple block~\cite{alexnet,Simonyan2014vgg,googlenet,andrew2017mobilenet} and the residual block~\cite{resnet}, the fact of whether the distillation happens before or after the ReLU constitutes the difference between our proposed method and other methods.
However, for networks using the pre-activation~\cite{He2016preact,Zagoruyko2016WRN}, the difference is larger.
Since there is no ReLU at the end of each block, our method has to find the ReLU in the next block.
In a structure like PyramidNet~\cite{Han2017pyramidnet,Sandler2018mobileV2}, our proposed method can reach the ReLU after the 1$\times$1 convolution layer.
Though the positioning strategy may be complicated according to the architecture, it has a significant influence on the performance.
Our new distillation position significantly improves the performance of the student as demonstrated in our experiments.

\subsection{Loss function}
\label{sec:loss_function}

\begin{figure}[t]
	\centering
    \includegraphics[width=\columnwidth]{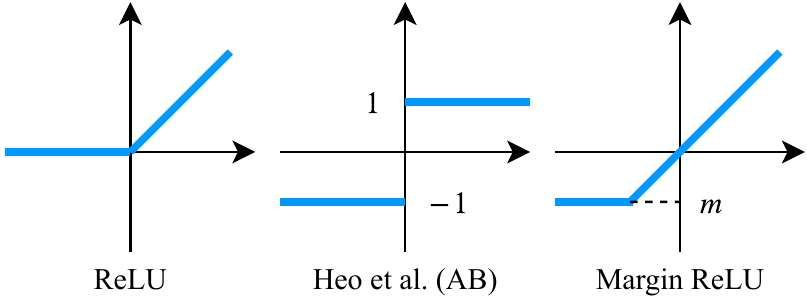}
  \caption{A comparison of the conventional ReLU, teacher transforms in Heo \textit{et al}.~\cite{heo2018knowledge} and our proposed method.}
\label{fig:mReLU}
\vspace{-0.2cm}
\end{figure}

Based on the format of section~\ref{section:preliminary}, we explain the teacher transform $T_t$, student transform $T_s$ and the distance function $d$ of our proposed method.
Since the feature values of teacher $\boldsymbol{F}_t$ are the values before ReLU, positive values have the information utilized by the teacher while negative values do not. %  the ones not used in the teacher network.
If a value in the teacher is positive, the student must produce the same value as in the teacher.
On the contrary, if a value of the teacher is negative, the student should produce a value less than zero to make same activation status of neurons.
Heo \textit{et al.}~\cite{heo2018knowledge} noted that a margin is required to make the student's value less than zero.
Thus, we propose a teacher transform that preserves positive values while giving a negative margin.
\begin{equation}
\sigma_{m} (x) = \mathrm{max}(x, m).
\label{equ:marginrelu}
\end{equation}
Here, $m$ is a margin value less than zero.
We name this function as margin ReLU.
Several types of teacher transforms are depicted in Fig.~\ref{fig:mReLU}.
Margin ReLU is designed to give a negative margin 
which is easier to follow than the negative value of the teacher.
Heo \textit{et al.} set the margin by an arbitrary scalar value, which does not reflect the weight values of the teacher.
In our proposed method, the margin value $m$ is defined as the channel-by-channel expectation value of the negative response, and the margin ReLU uses values that correspond to each channel of the input.
For a channel $\mathcal{C}$ and the $i$-th element of the teacher's feature $F^i_t$, the margin value of a channel $m_\mathcal{C}$ is set to an expectation value over all training images.
\begin{equation}
m_\mathcal{C} = E[F^i_t | F^i_t < 0, i \in \mathcal{C}].
\label{equ:marginrelu}
\end{equation}
The expectation value can be calculated directly in the training process, or it can be calculated using parameters of the previous batch normalization layer.
The margin ReLU $\sigma_{m_\mathcal{C}}(\cdot)$ is used as a teacher transform $T_t$ in our proposed method and produces the target feature value for the student network.
For the student transform, a regressor consisting of an 1 $\times$ 1 convolution layer~\cite{romero2014fitnets,heo2018knowledge} and a batch normalization layer is used.

We now explain our distance function $d$.
Our proposed method transfers the representation before ReLU.
Therefore, the distance function should be changed considering ReLU.
In the feature of the teacher, the positive responses are actually used for the network which implies that the positive responses of the teacher should be transferred by their exact values.
However, negative responses are not.
For a negative teacher response, if the student response is higher than the target value, it should be reduced,
but if the student response is lower than the target value, it doesn't need to be increased since negatives are equally blocked by ReLU regardless of their values.
For the feature representation of the teacher and student, $\boldsymbol{T}, \boldsymbol{S} \in \mathbb{R}^{W \times H \times C}$, let the $i$-th component of the tensor be $T_i, S_i \in \mathbb{R}$.
Our partial $L_2$ distance ($d_p$) is defined as
\begin{equation}
d_p(\boldsymbol{T}, \boldsymbol{S}) = \sum^{WHC}_{i} 
\begin{cases}
0 & \text{if } S_i \leq T_i \leq 0 \\ %T_{i} \leq 0, T_{i} > S_{i} \\
(T_{i} - S_{i})^2 & \text{otherwise}.
\end{cases}
  \label{equ:distance}
\end{equation}
where $\boldsymbol{T}$ is the position for the teacher's feature and $\boldsymbol{S}$ is the position for the student's feature.

Our proposed method uses margin ReLU $\sigma_{m_\mathcal{C}} (x)$ as a teacher transform $T_t$ and a regressor $r(\cdot)$ consisting of an 1$\times$1 convolution layer as a student transform $T_s$, and uses partial $L_2$ distance ($d_p$) as the distance function.
Distillation loss of the proposed method is:
\begin{equation}
    L_{distill} = d_p(\sigma_{m_\mathcal{C}}(\boldsymbol{F}_t), r(\boldsymbol{F}_s)).
\label{equ:loss}
\end{equation}
Our proposed method is conducted as \textit{continuous distillation} using the distillation loss $L_{distill}$.
Thus, the final loss function is the sum of distillation loss and task loss:
\begin{equation}
loss = L_{task} + \alpha L_{distill}.
\label{equ:contidistill2}
\end{equation}
The task loss refers to the loss specified by the task of a network.
The feature position for distillation is after the last block of one spatial size and before ReLU as depicted in Fig.~\ref{fig:position}.
In a network with a 32$\times$32 input, such as CIFAR~\cite{CIFAR}, there are three target layers, and in the case of ImageNet~\cite{ImageNet}, the number of target layers is four.

\input{Tables/table_CIFAR_100.tex}
\subsection{Batch normalization}
\label{sec:batchnorm}
We further investigate batch normalization in knowledge distillation.
Batch normalization~\cite{Ioffe2015batchnorm} is used in most recent network architectures to stabilize training.
A recent study on batch normalization~\cite{Ioffe2017batchrenorm} explains the difference between training mode and evaluation mode of batch normalization.
Each mode of batch-norm layer acts differently in the network.
Therefore, when performing knowledge distillation, it is necessary to consider whether to use the teacher in training mode or evaluation mode.
Typically, the feature of the student is normalized batch by batch.
Therefore, the feature from the teacher must be normalized in the same way.
In other words, the mode of the teacher's batch normalization layers should be training mode when distilling the information.
To do this, we attach a batch normalization layer after the 1$\times$1 convolutional layer and use it as a student transform and bring the knowledge from the teacher in training mode.
As a result, our proposed method achieves additional performance improvements.
This issue holds for all knowledge distillation methods including the proposed method.
We empirically analyze various knowledge distillation methods for batch normalization issues in Section~\ref{sec:batchnorm_exp}.

%% file: Tables/table_CIFAR_100.tex
\begin{table*}[t]
\begin{center}
\begin{tabular}{@{}ccccccc@{}}
\toprule
Setup & Compression type       & Teacher network      & Student network     & \begin{tabular}[c]{@{}c@{}}\# of params\\teacher\end{tabular} & \begin{tabular}[c]{@{}c@{}}\# of params\\student\end{tabular} & \begin{tabular}[c]{@{}c@{}}Compress\\ratio\end{tabular} \\ \midrule
(a)   & Depth                  & WideResNet 28-4      & WideResNet 16-4     & 5.87M        & 2.77M        & 47.2\%     \\
(b)   & Channel                & WideResNet 28-4      & WideResNet 28-2     & 5.87M        & 1.47M        & 25.0\%     \\
(c)   & Depth \& channel       & WideResNet 28-4      & WideResNet 16-2     & 5.87M        & 0.70M        & 11.9\%     \\
(d)   & Different architecture & WideResNet 28-4 & ResNet 56     & 5.87M       & 0.86M        & 14.7\%     \\
(e)   & Different architecture & PyramidNet-200 (240) & WideResNet 28-4     & 26.84M       & 5.87M        & 21.9\%     \\
(f)   & Different architecture & PyramidNet-200 (240) & PyramidNet-110 (84) & 26.84M       & 3.91M        & 14.6\%     \\ \bottomrule
\end{tabular}
\end{center}
\caption{Experiments settings with various network architectures on CIFAR-100. Network architecture is denoted as WideResNet (depth)-(channel multiplication) for Wide Residual Networks~\cite{Zagoruyko2016WRN} and PyramidNet-(depth) (channel factor) for PyramidNet~\cite{Han2017pyramidnet}.}
\label{table:cifarsetup}
\end{table*}

\begin{table*}[t]
\begin{center}
\begin{tabular}{@{}c|c|cccccccc@{}}
\toprule
Setup & Teacher & Baseline & KD~\cite{hinton2015distilling} & FitNets~\cite{romero2014fitnets}  & AT~\cite{zagoruyko2016paying}      & Jacobian~\cite{suraj2018jacobian} & FT~\cite{kim2018paraphrasing}      & AB~\cite{heo2018knowledge}      & Proposed \\ \midrule
(a)   & 21.09 & 22.72  & 21.69 & 21.85 & 22.07 & 22.18  & 21.72 & 21.36 & \textbf{20.89}  \\
(b)   & 21.09 & 24.88  & 23.43 & 23.94 & 23.80 & 23.70  & 23.41 & 23.19 & \textbf{21.98}  \\
(c)   & 21.09 & 27.32  & 26.47 & 26.30 & 26.56 & 26.71  & 25.91 & 26.02 & \textbf{24.08}  \\
(d)   & 21.09 & 27.68  & 26.76 & 26.35 & 26.66 & 26.60  & 26.20 & 26.04 & \textbf{24.44}  \\
(e)   & 15.57 & 21.09  & 20.97 & 22.16 & 19.28 & 20.59  & 19.04 & 20.46 & \textbf{17.80}  \\
(f)   & 15.57 & 22.58  & 21.68 & 23.79 & 19.93 & 23.49  & 19.53 & 20.89 & \textbf{18.89}  \\ \bottomrule
\end{tabular}
\end{center}
\caption{Performance of various knowledge distillation methods on CIFAR-100. Measurement is error rate (\%) of classification. lower is better. `Baseline' represents a result without distillation.}
\label{table:cifarperfomance}
\vspace{-0.3cm}
\end{table*}

%% file: 4_Experiments.tex
\section{Experiments}

We have evaluated the efficiency of our distillation method in several domains.
The first task is the classification problem which is a fundamental problem in machine learning.
As most of the other distillation methods have reported their performance under this domain, we also have compared our results to those of others.
The performance of knowledge distillation depends on which network architecture is used, how well the teacher performs and what kind of training scheme is used.
To control other factors and make a fair comparison, we reproduced the algorithms of other methods based on their codes and papers.
All experiments were implemented and evaluated on NAVER Smart Machine Learning (NSML)~\cite{nsml} platform with PyTorch~\cite{paszke2017pytorch}.

\subsection{CIFAR-100}
CIFAR-100~\cite{CIFAR} is the dataset that most knowledge distillation methods use to validate their performance.
Composed of 50,000 images with 100 classes, we use CIFAR-100 to compare various settings of all methods.
To be practically used in any task, knowledge distillation must be applicable to any network structure.
Therefore, we provide the experimental results of knowledge distillation using various structures of the teacher and student.
Table.~\ref{table:cifarsetup} shows the settings of each experiment such as the architecture used for each model, model size and compression rate.
Majority of our experiments utilize Wide Residual Network~\cite{Zagoruyko2016WRN} since the number of layers and the depth of each layer can be easily modified.
Distillation between different types of architectures has also been experimented with the setting (d), (e), (f).
In the case of (f), network names are similar, but the teacher is based on the bottleneck block and the student uses the basic block.
Note that (e) and (f) use a teacher network PyramidNet-200~\cite{Han2017pyramidnet} trained with the Mixup augmentation~\cite{zhang2018mixup}.
All models have been trained for 200 epochs with a learning rate of 0.1, multiplied by 0.1 at epoch 100 and 150.
To produce the best results from other methods, some algorithms~\cite{romero2014fitnets,kim2018paraphrasing,heo2018knowledge} are trained with an output distillation loss~\cite{hinton2015distilling} along with the feature distillation loss.
The rest of the algorithms have shown better results when trained without the output distillation loss.
The results of each method on every setting are presented in Table~\ref{table:cifarperfomance}.
Our proposed method outperforms the state-of-the-art in the settings of depth and channel compression (a), (b), (c) and different architecture (d), (e), (f).
Especially, in the setting of depth compression (a), the student network trained by our proposed method outperforms the teacher network.
The proposed method consistently shows a good performance regardless of the compression rate and even when distilling to different types of network architecture.
Note that the error rate of 17.8\% in (e) is better than any network reported in the paper of Wide Residual Network~\cite{Zagoruyko2016WRN}.
Therefore, our proposed method can be applied not only to small networks but also to large networks with high performance.

\subsection{ImageNet}
The image size of 32$\times$32 in CIFAR is not enough to represent real world images.
For this reason, we have conducted experiments on the ImageNet dataset~\cite{ImageNet} as well.
ImageNet includes images with an average size of 469$\times$387, which allows us to verify distillation performance in large images.
In this paper, we have used the dataset in ILSVRC 2012~\cite{ImageNet}.
This dataset consists of 1.2 million training images and 50 thousand validation images.
Images are cropped to the size of 224$\times$224 for training and evaluation.
The student network is trained for 100 epochs, and the learning rate begins at 0.1 multiplied by 0.1 at every 30th epoch.
For a fair comparison and a simple reproduction, we used the pre-trained model in the PyTorch~\cite{paszke2017pytorch} library as the teacher network.

The experiments have been conducted on two pairs of networks.
The first one is distillation from ResNet152~\cite{resnet} to ResNet50, and the second one is distillation from ResNet50 to MobileNet~\cite {andrew2017mobilenet}.
In this section, we present the results of three latest algorithms~\cite{zagoruyko2016paying, kim2018paraphrasing, heo2018knowledge}, which have shown the best results in the previous subsection.
The results are represented in Table~\ref{table:imagenet}.
Our proposed method shows a great improvement.
In particular, our method has made ResNet50 perform better than ResNet152, which is a remarkable achievement.
In addition, it has shown a considerable improvement in the recently proposed lightweight architecture, MobileNet.
In case of MobileNet, it is hard to reproduce the performance of the paper (29.4) because the training scheme, such as training epochs, is not reported.
Thus, we measured the performance in a standard setting.

\input{Tables/table_ImageNet.tex}
\subsection{Object detection}
\input{Tables/table_detection.tex}
\input{Tables/table_segmentation.tex}
In this section, we apply our method to other computer vision tasks.
The first one is object detection which is one of the most frequently used neural network techniques.
Since the purpose of distillation is to improve speed, we applied our proposed method on a high-speed detector, Single Shot Detector (SSD)~\cite{Liu2016SSD}.
Networks are trained on a mixture of VOC2007 and VOC2012~\cite{Everingham2015VOC} \textit{trainval} set, which are widely used in object detection.
The backbone network in all models is pre-trained using the ImageNet dataset.
Networks have been trained for 120k iterations with a batch size of 32.
To show the improvement of our method, we set the SSD trained with no distillation as the baseline, referred to as `Baseline' in Table~\ref{table:detection}.
SSD detector based on ResNet50~\cite{resnet} or VGG~\cite{Simonyan2014vgg} is used as the teacher network to examine the difference of performance according to the teacher architecture.
As the student networks, SSD based on ResNet18 and SSD lite based on Mobilenet~\cite{andrew2017mobilenet} have been used.

Detection performance has been evaluated in the VOC 2007 \textit{test} set and all results are presented in Table~\ref{table:detection}.
In the case of ResNet18, the performance improvement of ResNet18-T1 using ResNet teacher is larger than T2 using a VGG teacher.
Though both student architectures outperform the baseline, the distillation between similar structure shows a better quality of knowledge distillation.
In the case of MobileNet, our proposed method shows a constant performance improvement regardless of the type of the teacher.
Student models in all experiments have experienced improvements in performance and this implies that our method can be applied to any SSD-based object detector.

\subsection{Semantic segmentation}

In this section, we verify the performance of our proposed method on semantic segmentation.
Applying distillation on semantic segmentation is challenging since the output size is much larger than any other task.
We have selected the latest study, DeepLabV3+~\cite{Chen2018DeepLab} as our base model for semantic segmentation.
DeepLabV3+ based on ResNet101 has been used as the teacher network, and DeepLabV3+ based on MobileNetV2~\cite{Sandler2018mobileV2} and ResNet18~\cite{resnet} has been used as the student network.
Experiments have been performed on the PASCAL VOC 2012 segmentation~\cite{Everingham2015VOC} dataset.
We also use an augmentation of the dataset provided by the extra annotations in \cite{BharathICCV2011} as in the baseline paper~\cite{Chen2018DeepLab}.
All models have been trained for 50 epochs, and the learning rate schedule is the same as the baseline paper~\cite{Chen2018DeepLab}.
In similar fashion to our detection task, the student network is initialized to a network pre-trained on ImageNet without distillation.
Results are presented in Table~\ref{table:segmentation}.
Our proposed method significantly improves the performance of ResNet18 and MobilenetV2.
Taking MobileNetV2, in particular, our proposed method improves the performance by almost 3 points in mIoU and contributes to computation reduction of the segmentation algorithm.
We have shown that our proposed method can be applied to image classification, object detection and semantic segmentation.
Being able to be applied to many tasks without major changes is an advantage of feature distillation and shows that our proposed method has a wide range of applications.

\subsection{Analysis}

We analyze possible factors which would have lead to the performance improvement by our proposed method.
The first analysis is the output similarity between the teacher and the student learned by distillation.
By this, we verify how well our method forces the student to follow the teacher.
After that, we provide an ablation study of our proposed method.
We measure how much each component of our proposed method contributes to the performance.
Finally, we discuss about how the mode of batch normalization affects knowledge distillation, as mentioned in Section~\ref{sec:batchnorm}.
All experiments are based on setting (c) of Table~\ref{table:cifarsetup}.

\subsubsection{Teacher-student similarity}
\input{Tables/table_test_entropy.tex}
\input{Tables/table_batchnorm.tex}

KD~\cite{hinton2015distilling} forces the output of the student to be similar to output of the teacher.
The purpose of output distillation is quite intuitive, i.e., if
a student produces an output similar to that of the teacher, its performance will also be similar.
However, in the case of feature distillation, it is necessary to investigate how the output of the student changes.
To see how well the student mimics the teacher, we measure the similarity of the teacher's and student's output under a consistent setting.
On the test set of CIFAR-100, we measure the KL divergence between the teacher and student output.
The cross-entropy with the ground truth has also been measured since classification performance also contributes to the reduction of KL divergence.
Results are presented in Table~\ref{table:testentropy}.
Methods that apply distillation only in the early stage of the training, \textit{Initial distillation} (FitNets~\cite{romero2014fitnets}, AB~\cite{heo2018knowledge}), increase the KL divergence both with the teacher and ground-truth.
With this result, it is hard to say that the student networks of these methods are mimicking their teacher networks.
Meanwhile, distillation methods with \textit{continuous distillation} in Eq.~\ref{equ:contidistill2} (KD~\cite{hinton2015distilling}, AT~\cite{zagoruyko2016paying}, Jacobian~\cite{suraj2018jacobian}, FT~\cite{kim2018paraphrasing}) as well as our proposed method reduce the KL divergence, which implies that the similarity between the teacher and student is relatively high.
Specifically, our method shows a considerably high similarity compared to other \textit{continuous distillation} methods.
In other words, our proposed method trains the student to produce an output most similar to that of the teacher.
This similarity is one of the main reasons of improved performance of our proposed method.

\subsubsection{Ablation study}
Ablation experiments were conducted in which the ablation components were added one-by-one to measure their effects. 
The result is shown in Table~\ref{table:ablation}.
The baseline is a distillation method based on $L_2$ loss at end of block position. 
The version that uses the preReLU position (Section~\ref{sec:distill_position}) provides the greatest improvement because it is helpful to transfer the activation boundary effectively with both negative and positive values before ReLU.
The second improvement is achieved by the loss function (Section~\ref{sec:loss_function}), which prevents the transfer of useless and harmful negative values of less than a small negative margin. 
The batch-norm mode (Section~\ref{sec:batchnorm}) also contributes to the performance improvement.
In conclusion, a combination of all proposed components leads to a significant improvement in performance of the proposed method.

\subsubsection{Batch normalization}
\label{sec:batchnorm_exp}

In Section~\ref{sec:batchnorm}, we mentioned about the issue related to the mode of the batch normalization in knowledge distillation.
To investigate this, we measure the performance variation of knowledge distillation methods when differing the mode of the teacher's batch norm layer.
The experimental results are shown in Table~\ref{table:batchnorm}.
The distillation methods that use additional information other than feature (KD~\cite{hinton2015distilling}, Jacobian~\cite{suraj2018jacobian}) show marginal differences between each mode of batch normalization.
AT~\cite{zagoruyko2016paying}, which uses a diminished feature for distillation, has shown a better result in the evaluation mode.
However, methods that do not squeeze the feature (FitNets~\cite{romero2014fitnets}, FT~\cite{kim2018paraphrasing}, AB~\cite{heo2018knowledge}) consistently work better in the training mode.
Our method especially shows a substantial improvement when using the training mode.
Note that all experiments in previous sections exploit the better mode of the batch-norm layer as there is no mention about it in each paper.
In conclusion, an appropriate type of batch normalization should be carefully chosen in many distillation methods including ours.

\input{Tables/table_ablation.tex}

%% file: Tables/table_ImageNet.tex
\begin{table}[t]
\begin{center}
\begin{tabular}{@{}ccccc@{}}
\toprule
Network                   & \begin{tabular}[c]{@{}c@{}}\# of param\\(ratio)\end{tabular} & Method   & \begin{tabular}[c]{@{}c@{}}Top-1\\error(\%)\end{tabular}  & \begin{tabular}[c]{@{}c@{}}Top-5\\error(\%)\end{tabular}  \\ \midrule
ResNet152                 & 60.19M& Teacher  & 21.69 & 5.95 \\ \midrule
\multirow{6}{*}{ResNet50} & \multirow{6}{*}{\begin{tabular}[c]{@{}c@{}}25.56M\\(42.5\%)\end{tabular}}& Baseline & 23.72 & 6.97  \\
                          & & KD~\cite{hinton2015distilling}       & 22.85  & 6.55  \\
                          & & AT~\cite{zagoruyko2016paying}       & 22.75  & 6.35  \\
                          & & FT~\cite{kim2018paraphrasing}       & 22.80  & 6.49  \\
                          & & AB~\cite{heo2018knowledge}       &  23.47      &  6.94     \\
                          & & Proposed & \textbf{21.65} & \textbf{5.83} \\ \midrule \midrule
ResNet50                  &25.56M & Teacher  & 23.84  & 7.14 \\ \midrule
\multirow{6}{*}{MobileNet}& \multirow{6}{*}{\begin{tabular}[c]{@{}c@{}}4.23M\\(16.5\%)\end{tabular}} & Baseline & 31.13  & 11.24    \\
                          & & KD~\cite{hinton2015distilling}       & 31.42  & 11.02  \\
                          & & AT~\cite{zagoruyko2016paying}       &  30.44 & 10.67 \\
                          & & FT~\cite{kim2018paraphrasing}       & 30.12  & 10.50 \\
                          & & AB~\cite{heo2018knowledge}       &  31.11  & 11.29 \\
                          & & Proposed & \textbf{28.75}   & \textbf{9.66}  \\ \bottomrule
\end{tabular}
\end{center}
\caption{Results on ILSVRC 2012 validation set. Networks are trained and evaluated in 224$\times$224 size with single-crop. `Baseline' represents a result without distillation.}
\label{table:imagenet}
\vspace{-0.2cm}
\end{table}

%% file: Tables/table_detection.tex
\begin{table}[t]
\begin{center}
\begin{tabular}{@{}cccc@{}}
\toprule
Network                               & \# of params           & Method  & mAP(\%)    \\ \midrule
ResNet50-SSD                       & 36.7M                  & Teacher (T1)  & 76.79 \\ 
VGG-SSD                            & 26.3M                  & Teacher (T2) & 77.50 \\ \midrule
\multirow{3}{*}{ResNet18-SSD}      & \multirow{3}{*}{20.0M} & Baseline & 71.61 \\
                                      &                        & Proposed-T1 & \textbf{73.08} \\
                                      &                        & Proposed-T2 & 72.38 \\ \midrule
\multirow{3}{*}{\begin{tabular}[c]{@{}c@{}}MobileNet\\-SSD lite\end{tabular}} & \multirow{3}{*}{6.5M}  & Baseline & 67.58 \\
                                      &                        & Proposed-T1 & \textbf{68.54} \\
                                      &                        & Proposed-T2 & 68.45 \\ \bottomrule
\end{tabular}
\end{center}
\caption{Object detection results of SSD300~\cite{Liu2016SSD} in PASCAL VOC2007 \textit{test} set~\cite{Everingham2015VOC}. Results are described in mean Average Precision (mAP). Higher is better. %`Baseline' represents a result without distillation.
}
\label{table:detection}
\vspace{-0.3cm}
\end{table}

%% file: Tables/table_segmentation.tex
\begin{table}[t]
\begin{center}
\begin{tabular}{@{}cccc@{}}
\toprule
Backbone                               & \# of params           & Method   & mIoU    \\ \midrule
ResNet101                       & 59.3M                  & Teacher  & 77.39 \\ \midrule
\multirow{2}{*}{ResNet18}      & \multirow{2}{*}{\begin{tabular}[c]{@{}c@{}}16.6M\\(28.0\%)\end{tabular}} & Baseline & 71.79 \\
                                     &                        & Proposed & \textbf{73.24} \\ \midrule
\multirow{2}{*}{MobileNetV2}      & \multirow{2}{*}{\begin{tabular}[c]{@{}c@{}}5.8M\\(9.8\%)\end{tabular}} & Baseline & 68.44 \\
                                     &                        & Proposed & \textbf{71.36}  \\ \bottomrule                                     
\end{tabular}
\end{center}
\caption{Semantic segmentation based on DeepLabV3+~\cite{Chen2018DeepLab} on the PASCAL VOC 2012 \textit{test} set~\cite{Everingham2015VOC}. Measurement of performance is mean Intersection over Union (mIoU).}
\label{table:segmentation}
\vspace{-0.2cm}
\end{table}

%% file: Tables/table_test_entropy.tex
\begin{table}[t]
\begin{center}
\begin{tabular}{@{}lccc@{}}
\toprule
Method   & \begin{tabular}[c]{@{}c@{}}KL divergence\\ with teacher\end{tabular} & \begin{tabular}[c]{@{}c@{}}Cross-entropy\\ with GT\end{tabular} & \begin{tabular}[c]{@{}c@{}}Error\\rate(\%)\end{tabular}\\ \midrule
Baseline & 0.7318                                                               & 1.0741                                                          & 27.32    \\
KD~\cite{hinton2015distilling}       & 0.7064                                                               & 1.0758                                                          & 26.47    \\
FitNets~\cite{romero2014fitnets}   & 0.7993                                                               & 1.1585                                                          & 26.30    \\
AT~\cite{zagoruyko2016paying}       & 0.7047                                                               & 1.0303                                                          & 26.56    \\
Jacobian~\cite{suraj2018jacobian} & 0.7122                                                               & 1.0495                                                          & 26.71    \\
FT~\cite{kim2018paraphrasing}       & 0.6872                                                               & 1.0561                                                          & 25.91    \\
AB~\cite{heo2018knowledge}       & 0.7555                                                               & 1.1197                                                          & 26.02    \\
Proposed & \textbf{0.5723}                                                               & \textbf{0.9585}                                                          & \textbf{24.08}    \\ \bottomrule

\end{tabular}
\end{center}
\caption{Output similarity analysis between teacher and student on test set of CIFAR-100.}
\label{table:testentropy}
\vspace{-0.2cm}
\end{table}

%% file: Tables/table_batchnorm.tex
\begin{table*}[t]
\begin{center}
\begin{tabular}{@{}cccccccc@{}}
\toprule
   Mode of batch-norm       & KD~\cite{hinton2015distilling}     & FitNets~\cite{romero2014fitnets} & AT~\cite{zagoruyko2016paying}     & Jacobian~\cite{suraj2018jacobian} & FT~\cite{kim2018paraphrasing}     & AB~\cite{heo2018knowledge}     & Proposed \\ \midrule
Training mode & 26.47   & 26.61 & 26.56 & 26.71         & 25.91 & 26.02 & 24.08  \\
Evaluation mode  & 26.45   & 26.92 & 26.42 & 26.75         & 26.15 & 26.36 & 24.54  \\ \bottomrule
\end{tabular}
\end{center}
\caption{Analysis of the mode of batch normalization in teacher network on CIFAR-100. Table shows error rate(\%). The first row shows the results of teacher's batch-norm in training mode while the second row shows the results of using the batch-norm in evaluation mode.}
\label{table:batchnorm}
\vspace{-0.2cm}
\end{table*}

%% file: Tables/table_ablation.tex
\begin{table}[t]
\begin{center}
\begin{tabular}{@{}ccccc@{}}
\toprule
& Baseline & + \begin{tabular}[c]{@{}c@{}}Position\\ (Sec.\ref{sec:distill_position})\end{tabular}  & + \begin{tabular}[c]{@{}c@{}}BN\\ (Sec.\ref{sec:batchnorm})\end{tabular} & +  \begin{tabular}[c]{@{}c@{}} loss \\(Sec.\ref{sec:loss_function}) \end{tabular}  \\ \midrule
Error & 26.37 &  24.81 & 24.68 & 24.08 \\ 
Diff & -  &  -1.56 & -0.13 & -0.60 \\ \bottomrule
\end{tabular}
\end{center}
\caption{Ablation study of proposed method. The results are presented in the form of error rate (\%).}
\label{table:ablation}
\vspace{-0.3cm}
\end{table}

%% file: 5_Conclusion.tex
\section{Conclusion}

We propose a new knowledge distillation method along with several investigations about various aspects of the existing feature distillation methods.
We have discovered the effectiveness of pre-ReLU location and proposed a new loss function to improve the performance of feature distillation. 
The new loss function consists of a teacher transform (margin ReLU) and a new distance function (partial $L_2$) and enables an effective feature distillation at pre-ReLU location.
We have also investigated about the mode of batch normalization in teacher network and achieved additional performance improvements.
Through experiments, we examined the performance of the proposed method using various networks in various tasks, and proved that the proposed method substantially outperforms the state-of-the-arts of feature distillation.

%% file: Appendix.tex
\section*{Appendix}

%%%%%%%%% BODY TEXT
\subsection*{A. margin evaluation}

We calculated margin of each channel and use margin ReLU with channel margin $m_\mathcal{C}$.
The margin is the expectation of the negative value of the feature, which can be obtained directly during training or using a batch normalize layer.
We explains how to obtain the margin value using a batch norm layer.
For a channel $\mathcal{C}$ and the $i$-th element of teacher's feature $F^i_t$, the margin value of a channel $m_\mathcal{C}$ is set to an expected value over training images.

\begin{equation}
m_\mathcal{C} = E[F^i_t | F^i_t < 0, i \in \mathcal{C}].
\end{equation}
In general, we can't know the distribution of $F^i_t$, so expectation must be obtained through average operation over training process.
However, when a batch-norm layer prior to ReLU, the batch-norm layer determines the distribution of feature $F^i_t$ in a batch.
Batch norm layer normalizes the feature for each channel to a gaussian distribution with a specific mean $\mu$ and variance $\sigma$.
In other words,

\begin{equation}
F_t^i \sim \mathcal{N} (\mu, \sigma).
\end{equation}
The value of mean and variance ($\mu, \sigma$) of each channel correspond to the parameters ($\beta, \gamma$) of the batch-norm layer.
So, it can be obtained by analyzing the teacher network.
Using the distribution of $F^i_t$, we can directly calculate the margin value.

\begin{equation}
m_\mathcal{C} = \frac{1}{Z} \int_{-\infty}^0 \frac{x}{\sqrt{2 \pi \sigma}} e^{-\frac{(x-\mu)^2}{2 \sigma^2}} dx
\end{equation}
The expectation can be obtained from integration using pdf of gaussian distribution, where the range is smaller than zero.
The result of the integration can be expressed in simple form using the cdf function $\Phi(\cdot)$ of normal distribution.

\begin{equation}
m_\mathcal{C} = \mu - \frac{\sigma e^{-\mu^2 / 2 \sigma^2}}{\sqrt{2 \pi} \Phi(-\mu / \sigma)}
\label{equ:final}
\end{equation}
Using Eq.~\ref{equ:final}, the proposed method obtains channel-wise margin value without sampling and averaging on training process.
In the experiment of the paper, if the ReLU is followed by batch normalize, the margin is obtained by using Eq.~\ref{equ:final}. Otherwise, the expectation is obtained from average operation on training process.

\subsection*{B. implementation details}
Features for distillation are selected just before down-sampling layers, which total three layers for CIFAR and four layers for ImageNet.
In the loss function of our method in Eq.~\ref{equ:loss}, we sum the values in the entire layer rather than averaging them.
When one moves from the top layer to the bottom layer, the total size of the feature is increased by twice the amount as spatial resolution increases.
Therefore, the loss is divided in half accordingly.
$\alpha$ in Eq.~\ref{equ:contidistill2} is $10^{-3}$ for CIFAR~\cite{CIFAR}, $10^{-4}$ for ImageNet~\cite{ImageNet} and detection, and $10^{-5}$ for segmentation.
For CIFAR, we used a batch size of 128 for (a) to (d) and a batch size of 64 for (e) and (f).
% Every experiment on CIFAR has been performed 5 times with the median value reported.
% For ImageNet, batch size was 256.
Detection has an extra layer behind the backbone and all extra layers were also used for distillation.
Detector were trained over a 120k iteration with a batch size of 32.
The learning rate started at $10^{-3}$ and was multiplied by 0.1 at iteration 80k and 100k.
In the case of segmentation, we use an additional distillation layer at the \textit{atrous spatial pyramid pooling} and a layer just before output layer.
Output stride was set to 16 and all dropout layers are not used for distillation.
When using a pre-trained network, we initialized the student transform at the start of training.
Initialization proceeded for 500 iteration for detection and 1 epoch for segmentation.

\subsection*{C. additional experiments}

We measured the performance of other distillation methods at our preReLU position.
We conducted this experiment in setting (c) of Table~\ref{table:cifarsetup}.
As shown in the Table~\ref{table:preReLU}, the preReLU improves the performance of most algorithms.

\begin{table}[h]
\centering
\footnotesize
\begin{tabular}{@{}ccccccc@{}}
\toprule
Position     & FitNets & AT  & Jacobian & FT  & AB    & Proposed   \\ \midrule
Block & 26.30 & 26.42 & 26.71  & 25.91 & -     & -         \\
preReLU      & 26.22 & 26.45 & 26.27  & 25.11 & 26.02 & 24.08 \\ \bottomrule
\end{tabular}
\vspace{0.2cm}
\caption{Performance of other distillation methods in preReLU.}
\label{table:preReLU}
\end{table}
We also measured performance of proposed method in a single-layer setting.
As shown in the Table~\ref{table:multi-layer}, single-layer version is not significantly different from the multi-layer version, which implies that our method outperforms the existing methods in any settings.
\begin{table}[h]
\centering
\footnotesize
\begin{tabular}{@{}ccccccc@{}}
\toprule
Setup (in Table.~\ref{table:cifarsetup})       & (a)     & (b)     & (c)     & (d)     & (e)    & (f) \\ \midrule
Multi-layer  & 20.89 & 21.98 & 24.08 & 24.44 & 17.80 & 18.89 \\
Single-layer & 20.90 & 22.03 & 24.14 & 24.78 & 18.17 & 18.99 \\ \bottomrule
\end{tabular}
\vspace{0.2cm}
\caption{Comparison between single-layer and multi-layer implementation of proposed method.}
\label{table:multi-layer}
\end{table}